\documentclass{preprint}

\usepackage{array}      
\usepackage{xspace}
\usepackage{amsmath,amssymb,amsfonts}
\usepackage{booktabs}
\usepackage{graphicx}
\usepackage{tabularx}
\usepackage{xcolor}
\usepackage{tabularx}
\usepackage{spverbatim}
\usepackage{mdframed}
\usepackage{colortbl}
\usepackage[numbers,comma,sort&compress]{natbib} 
\usepackage{hyperref}

\newcommand{\redose}{\textsc{ReDose}\xspace}

\title{Curation and Extraction of Drug-Related Entities from Reddit Platform}

\author[1,$\dagger$]{Zewei Wang}
\author[1,2,$\dagger$]{Zihan Xu}
\author[1]{Yishu Wei}
\author[3,*]{Michael Chary}
\author[1,*]{Yifan Peng}
\affil[1]{Population Health Sciences,
Weill Cornell Medicine,
New York City, USA}
\affil[2]{School of Computing and Information Systems, University of Melbourne,
Melbourne, Australia}
\affil[3]{Emergency Medicine,
Weill Cornell Medicine,
New York City, USA}

\affil[$\dagger$]{These authors contributed equally to this work.}
\affil[*]{Corresponding author(s). Email(s): \url{mic9189@med.cornell.edu}, \url{yip4002@med.cornell.edu}}

\newcounter{prompt}
\newenvironment{prompt}[1][]{
\refstepcounter{prompt}
\begin{mdframed}[
innertopmargin=0pt, 
innerleftmargin=2pt, 
innerrightmargin=2pt,
frametitle={Box \theprompt. #1},
frametitlefont=\footnotesize,
frametitlerule=true,
backgroundcolor=gray!10]%
\setlength{\parindent}{0pt}%
\setlength{\parskip}{.5em}%
\scriptsize\ttfamily\hyphenchar\font=`\-\spaceskip=.5em plus .5em\xspaceskip=.5em%
}
{%
\par%
\end{mdframed}%
}

\begin{document}

\maketitle

\begin{abstract}
Physicians learn primarily about illicit drugs from clinical overdose cases, limiting their understanding of real-world usage. Meanwhile, drug users share first-hand experiences online, offering insights into dosage and effects of drugs. To bridge this gap, we introduce \redose (REddit Drug
DOSe and Effect), a dataset of 6,435 Reddit posts on substance use. A board-certified toxicologist primarily annotated both the training and test sets, while two medical science students contributed to the test set, labeling DRUG, DOSE, and EFFECT entities. We benchmarked 6,267 annotations using BERT-based, large language model (LLM)-based, and Retrieval-Augmented Generation (RAG) models. BiomedBERT achieved an F1-score of 0.843 for DRUG, while Llama-3 70B outperformed GPT-4 (F1 = 0.79 vs. 0.72). EFFECT extraction remains challenging, with GPT-4 achieving a recall of 0.41. \redose captures patient-curated narratives to advance medical data extraction from social media.

\end{abstract}


\section{Introduction}

The epidemiology of substance use has transformed from single-substance use to polysubstance use, and the inventory of possible drugs has expanded from a handful to a dizzying pantheon of novel psychoactive substances (NPS). NPS include novel synthetic opioids, hallucinogenic stimulants, and designer benzodiazepines. According to the United Nations Office of Drug Control, the number of known NPS increased from 251 in 2012 to 780 by 2016~\cite{Tettey2017}. These substances emerge and disappear too rapidly for federal surveys to track or routine toxicology screens to detect, necessitating the development of improved detection methods. People frequently discuss substance use on social media~\cite{graves2018opioid, graves, pandrekar2018social}, where a great proportion of the discussions is about NPS use~\cite{bowen2019increases, wright2021detection}. It has been previously demonstrated that online discussions on Twitter/X about opioid use can be used to predict real-world use in the next 30 days~\cite{Chary2017}. In addition to previous work on dose-response information extraction from online platforms such as YouTube comments~\cite{Chary2014} and online bulletin boards~\cite{abdelati2023sublethal}, Reddit is also frequently cited as a rich source of information on various medical topics~\cite{spadaro2022, graves}. Previous studies have shown that using social media text is a valid method for tracking how people use specific substances. However, less research has focused on identifying which new substances are emerging. This gap creates an opportunity to determine whether analyzing online commentaries can help identify problematic new substances before they cause severe public health issues.

Natural Language Processing (NLP) techniques are well-suited to processing large volumes of unstructured text. The current key challenge is the lack of standardized, curated datasets for benchmarking extraction methods. Existing Named Entity Recognition (NER) datasets from social media are only annotated on a few substance entities, but none of them touch on effects or doses~\cite{1490164, graves, spadaro2022, li2016, leas2020cbd, ge2024reddit}. Open-source clinical NLP datasets are annotated with drug dosages, but they are primarily based on structured physician narratives~\cite{1490164}. In contrast, the online commentaries do not strictly use clinical terminology to avoid sensitive checks from online platforms. This poses a significant challenge for developing models that transfer well between physician notes and textual data from social media.

To address this barrier, we introduce \redose (REddit Drug DOSe and Effect), a dataset of 6,435 unique documents collected from 7 drug-related subreddits. Each document within \redose has been annotated with three entities: the drugs mentioned, their reported doses, and the reported effects. We chose these three entities because establishing a dose-effect relationship is a cornerstone of pharmacology. For many substances described online, there is no other feasible data source for this purpose (Table~\ref{tab:datasets}). Each entry in \redose includes an unprocessed document, its annotated version, and a timestamp. All identifiable and protected health information has been removed. Our goal with \redose is to enhance the ability of NLP models to extract clinically relevant information about emerging substances, thereby informing clinical practice and public health guidelines. To our knowledge, \redose is the first such dataset collected from online platforms with such detailed annotations, including three attributes. 

\begin{table*}[t]
\rowcolors{2}{}{lightgray!35}
\caption{Summary of Datasets in Related Studies}
\label{tab:datasets}
\centering
\small
\begin{tabularx}{\textwidth}{l>{\hsize=1.4\hsize\linewidth=\hsize\raggedright\arraybackslash}X>{\hsize=.6\hsize\linewidth=\hsize\raggedright\arraybackslash}Xcc}
\toprule
\textbf{Dataset Name}
& \textbf{{Source}} & \textbf{{Annotated Drugs}}
& \textbf{{Open source}} & \textbf{{Social media}}\\
\midrule
Graves et al.~\cite{graves}
& Public posts from the /r/suboxone subreddit of Reddit
& Buprenorphine--naloxone & $\times$ & $\checkmark$\\
Leas et al.~\cite{leas2020cbd}
& r/CBD (2014--2019) & Cannabidiol & $\times$ & $\checkmark$\\
Spadaro et al.~\cite{spadaro2022}
& r/fentanyl, r/heroin, r/microdosing, r/opiates, r/OpiatesRecovery, r/OurOverUsedVeins, r/suboxone
& Buprenorphine and fentanyl & $\times$ & $\checkmark$\\
BC5CDR~\cite{li2016} & PubMed Articles & Chemicals & $\checkmark$ & $\times$\\
ADE~\cite{1490164} & MIMIC--III clinical care database & Prescription Medication & $\checkmark$ & $\times$\\
Reddit--Impacts~\cite{ge2024reddit} & Opioid--related subreddits on Reddit & None (impact only) & $\times$ & $\checkmark$\\
\textsc{ReDoSE} & r/fentanyl, r/heroin, r/microdosing, r/opiates, r/OpiatesRecovery, r/OurOverUsedVeins, r/suboxone
& All chemicals prone to be abused & $\checkmark$ & $\checkmark$\\
\bottomrule
\end{tabularx}
\end{table*}

We present benchmark results using BERT-based models, one-shot prompting Large Language Models (LLMs), and Retrieval-Augmented Generation (RAG)-based LLMs. In traditional one-shot or few-shot prompting, the same example is used for all inputs, which limits the impact of examples because their semantics may differ significantly from the input. To address this limitation, we developed a retrieval-based method to extract the most similar examples from the training dataset and append them to the prompt. This approach significantly improved the recall rate for DRUG extraction. In comparing BERT with LLMs, we found that while LLMs may be easier to implement, their performance does not yet match that of the fine-tuned BERT. The performance difference is most significant in the metrics on the EFFECT entity.  

In summary, this study provides several key contributions. (1) We introduce \redose, a comprehensive dataset of online commentaries about drug use. This dataset comprises 6,435 documents and 6,267 DRUG, DOSE, or EFFECT entities, with a high inter-annotator agreement score of 0.75. (2) We provide benchmark results for a fair comparison between BERT and LLMs. (3) We conduct a deep analysis of the differences across these models, exploring how the inherent large knowledge databases in LLMs may yield performance comparable to the supervised training results of BERT-based models. This analysis helps understand the strengths and limitations of each approach in handling complex medical NER tasks.
\section{Related Work}
\subsection{Biomedical Named Entity Recognition (NER) Dataset}

There have been abundant datasets designed for medication-related entity recognition. We collect the most relevant datasets and present their annotated attributes in Table~\ref{tab:datasets}. We highlight some major limitations of the existing datasets in the following discussion. 

First, some datasets extend beyond drugs to general medication. While most drugs have the potential to be abused, studies have shown that certain drugs, such as depressants, opioids or morphine derivatives, and nerve stimulants, appear to be more addictive than others \cite{drug}. \redose is a more focused dataset compared to a general medication-based dataset and is better suited to research focused on substance abuse. In addition, since adverse events are prevalent in \redose due to frequent overdosing, it can also serve as a complementary dataset in research related to adverse drug events.

Most existing datasets extracted from Reddit are not open-sourced, which limits their reproducibility and reuse. Meanwhile, widely used open datasets, such as n2c2, mainly rely on reports written by medical professionals that use standard clinical terminologies. By comparison, \redose introduces new terminology used by broader populations, helping physicians to familiarize themselves with the synonyms of the drugs. 

Moreover, some datasets only investigate a single substance. For example, in the work by Graves et al.~\cite{graves}, Reddit posts from the \textit{/r/suboxone} subreddit were mined to study user discussions around a specific drug (Suboxone), with a focus on its symptoms and usage patterns. While insightful, that approach was narrowly scoped to one substance and lacks coverage of the broader landscape of multiple substances across diverse user groups. By contrast, \redose expanded the source to 7 related subreddits spanning multiple drugs. This would allow medical researchers to have a wider perspective of commonly abused drugs. 

With regards to NLP techniques and substance abuse, Spadaro et al.~\cite{spadaro2022} investigated discussions of precipitated opioid withdrawal (POW) in the context of fentanyl and buprenorphine induction by analyzing 267,136 posts from seven opioid-related subreddits between 2012 and 2021. Using a combination of keyword searches and NLP filtering, they identified and thematically analyzed several hundred posts specifically referencing POW and microdosing (Bernese method). While their approach yielded valuable insights into community experiences, it was limited by its reliance on keyword-based retrieval, which may have excluded relevant discussions in other communities—potentially introducing selection bias and constraining the generalizability of their findings. Henry et al.~\cite{1490164} focused on the 2018 National NLP Clinical Challenges shared task, which aimed to extract Adverse Drug Events (ADEs) from clinical records. The task evaluated three main areas: concept extraction, relation classification, and end-to-end systems. The study employed deep learning-based methods, specifically BiLSTM-CRF models, and achieved high performance across various areas. However, BiLSTM-CRF models faced significant challenges in identifying ADEs and reason concepts because they require inference across multiple sentences or paragraphs. Symptoms or reactions may be implied rather than explicitly stated, causing difficulty for models that excel to local sequence labeling but struggle with long-range dependencies. 

In a recent dataset, `Reddit Impact', Ge et al.~\cite{ge2024reddit} analyzed clinical and social impacts of Substance Use Disorders (SUDs) using data from Reddit. It introduced the Reddit-Impacts dataset, derived from posts in fourteen opioid-related subreddits, aiming to capture the clinical and social impacts of substance use as reported by individuals discussing their personal experiences. The researchers adopted NLP techniques, including BERT, RoBERTa, DANN, and GPT-3.5, to automatically identify and classify these impacts. Despite the dataset's value in highlighting real-world impacts of SUDs, its limitations included the sparsity of annotated impacts, which instead relied on the LLM's judgment. This could lead to unreliable annotations, as language models struggle to accurately extract impacts from SUDs that contain slang. Potential selection bias may arise from focusing on specific subreddits and failing to accurately represent the broader population. 

Compared with other studies, our study features a higher standard of annotation, including the involvement of a medical toxicology specialist. Additionally, to ensure accuracy, additional annotators were brought in to annotate the documents in the validation dataset. Thus, we believe \redose covers a broader range of drugs with professional annotations, which makes it a stronger candidate for medical research. 

\subsection{Models on medical NER tasks}

\textbf{Large Language Models:} Since the era of LLMs, much work has been conducted on how they can be used in the medical field. Recent works by Li et al.~\cite{li2016} investigated the performance of various LLMs on medical NER. GPT-4 achieved satisfactory F1 scores, with models like PromptNER and GPT-NER achieving F1 scores above 90\% on the BC5CDR and NCBI datasets.
Ashok and Lipton~\cite{ashok2023promptner} introduced PromptNER, which used a chain-of-thought approach to improve named entity recognition by generating a logical sequence of steps to identify entities in text. Wang et al.~\cite{wang-etal-2025-gpt} introduced GPT-NER by appending special tokens and adding a self-verification strategy. 

Another work by Hu et al.~\cite{hu2024improving} evaluated the use of GPT-3.5 and GPT-4 for clinical NER tasks, focusing on datasets from MTSamples and VAERS. By employing a structured prompt engineering framework, the models demonstrated improved performance in extracting medical problems, treatments, and tests, as well as in identifying adverse events related to nervous system disorders. Despite these improvements, the GPT models still lag behind BioClinicalBERT, which had superior performance on both datasets. The study highlighted the potential of GPT models for clinical NER tasks, but also underscored the need for further refinement and the development of better evaluation metrics. All datasets and codes are publicly available, promoting further research and development in this area.

\textbf{Small Language Models:}
Small Language Models (SLMs) demonstrated greater sensitivity to the amount of training data, with performance improving significantly as the amount of data increased. For example, models like W-PROCER \cite{li2023wprocer} and MetaNER \cite{chen-etal-2023-learning} performed better on the 5-shot datasets than on the 1-shot datasets. However, SLMs struggled with fewer annotations and lacked the robustness seen in LLMs, particularly in scenarios with limited training data.

\section{Materials and Methods}
\subsection{Data source}\label{AA}
The documents in \redose were collected from seven subreddits detailed in Table~\ref{tab:subreddits}. These subreddits were chosen since previous smaller studies have demonstrated their richness and validity~\cite{graves, spadaro2022, ge2024reddit}. We first wrote custom programs using \textit{praw}~\cite{B2012}, a widely used Python wrapper for the Reddit API, which allows authenticated programmatic access to posts, comments, and metadata. This enables the extraction of the text of each post and its timestamp from these subreddits. To protect user privacy and ensure data integrity, metadata such as explicit usernames, geolocation references, or external links were discarded at the point of collection. Duplicate posts were also excluded. In other words, if the text was posted in two different subreddits, we removed all but one mention in the amalgamated dataset to prevent overrepresentation of repeated content. 

\begin{table}[t]
\rowcolors{2}{}{lightgray!35}
\caption{Summary of the 7 Subreddits}
\label{tab:subreddits}
\centering
\small
\begin{tabularx}{\textwidth}{lXr}
\toprule
\textbf{Subreddit} & \textbf{Description} & \textbf{\#Posts} \\
\midrule
r/fentanyl & Subreddit for the discussion of fentanyl and its many analogues. We focus on harm reduction, and hope to dispel some common myths about these substances. & 10,816 \\
r/heroin & For the junkies and ex-junkies of Reddit: A subreddit for all things heroin. Topics include heroin appreciation, harm reduction, withdrawal, recovery, lifestyle discussions, and more. & 31,322 \\
r/microdosing & This is a community for discussion pertaining to microdosing research, experiments, regimens, and experiences. The most probable candidates for microdosing are psychedelics, but we encourage dialogue on the effects of any drugs at sub-threshold dosage. No sourcing of drugs allowed! Please have a look at the r/microdosing Sidebar. & 42,766 \\
r/opiates & Discussion of all things related to the narcotics known as opiates, from harm-reduction to pharmacology. We are participating in the blackout! We will not be adding approved users to the community as we will return to normal functions tomorrow & 62,138 \\
r/OpiatesRecovery & You are not alone in this fight.
We are a group of people dedicated to helping each other stop and stay stopped. & 76,888 \\
r/OurOverUsedVeins & For those of us with an appreciation for heroin, to soothe, help, and give information on the subject of heroin use, and getting past our heroin addictions. Also, share your true stories! As heroin addicts, we've been in more terrifying situations than most people. OurOverUsedVeins wants to hear about them! & 79,581 \\
r/suboxone & For any and all discussion about the medication Suboxone (buprenorphine/naloxone). We hope to provide accurate, evidence-based information about Suboxone, resources for those interested in starting Suboxone, and support for those currently taking Suboxone. Whether you use Suboxone short-term or long-term; for Medication Assisted Treatment, pain management, or recreationally; are interested in starting Suboxone or just have someone in your life who takes Suboxone, you are welcome here! & 97,551 \\ 
\bottomrule
\end{tabularx}
\end{table}

\subsection{Annotation Process}
\begin{figure}[t]
\centering
\includegraphics[width=.7\columnwidth]{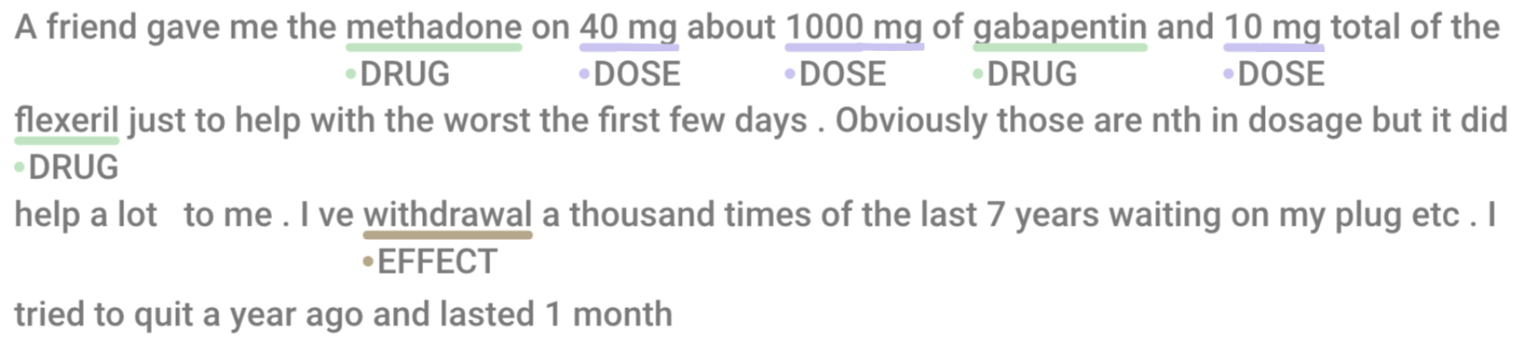}
\caption{A sample annotation with three types of entities.}
\label{fig:y equals x}
\end{figure}

Each document in \redose, including training and testing datasets, was annotated with three entities: the drug, its dose, and the reported effect, by a board-certified and clinically active medical toxicologist. A sample annotation with three types of entities was presented in Figure \ref{fig:y equals x}. Additionally, two medical science students independently annotated the validation dataset for the same entities. ``DRUG'' represents a drug, vitamin, or herb, but not a neurotransmitter, which is termed a xenobiotic in pharmacology and toxicology. ``DOSE'' indicates the quantity and units of the substance taken. In our annotation scheme, an $n$-gram of any size could be annotated. ``EFFECT'' represents a change in the physical or mental state that the writer attributed to the substance. For example, ``SEA \#4", ``black tar heroin", and ``cocaine" all received the label DRUG. We chose this approach to capture the same concept written in different ways. Online commentaries often use periphrastic or obfuscatory constructions. Spacing between words is frequently irregular: ``SEA\#4" and ``SEA \#4" occur with almost equal frequency. One goal of developing an NER module is to feed its output into a named entity linker. While ``heroin", ``SEA \#4", and ``black tar heroin" may all refer to the substance heroin, they are distinguished as different drug variants during the linking process, since ``SEA \#4" (southeast Asia \#4) is more potent and ``black tar heroin" is the formulation most associated with wound infections. 

\subsection{Validation of annotations}

To better quantify the agreement between two annotators, we employed a more suitable inter-annotator agreement (IAA) calculation method. The traditional IAA uses a binary measurement, which classifies each rater as fully agreeing or fully disagreeing with each document. This metric is less informative for documents that contain multiple entities. Only by analyzing the inter-rater agreement entity by entity can the traditional IAA calculation be effective. However, it would not assess how well \redose can train models for the intended use case. Thus, we adopted the method described by Jarrar et al.~\cite{jarrar2022wojood} for our needs: 
\begin{align}
   \kappa  &  =\frac{P_o - P_e}{1 - P_e} \label{eq:cohens-kappa}\\
   P_e  & =\frac{1}{N^2} \sum_{T} n_{T,1} \times n_{T,2}.
\end{align}
In this formula, $P_o$ refers to the observed agreement between annotators. $P_e$ refers to an expected agreement. $N$ represents the total number of annotations in the dataset, $T$  the number of different labels or categories in the label set, $n_{T,i}$ the number of times the annotator $i$ assigned the label $T$. 

\subsection{Developing the BERT-based models}
We first developed three BERT-based models: BaseBERT~\cite{Devlin2019-vb}, BioBERT~\cite{Lee2019}, and BiomedBERT~\cite{pubmedbert}. These models were pre-trained on general corpora, medical knowledge, and biomedical data, respectively, and then fine-tuned on the \redose training set. We excluded sentences that contain no entities from the training set. Since most articles exceed the token limit of BERT (512), we applied sentence tokenization using spaCy. The experiment was conducted on a single server equipped with two A6000 GPUs, each offering 48 GB of memory. The hyperparameters used in the experiment include 5 training epochs, a learning rate of 3e-5, a batch size of 8, and the default AdamW optimizer. 

To assess whether additional neural network layers could enhance performance, we incorporated two models that combine a conditional random field (CRF) and long short-term memory (LSTM) layers on top of the better-performing BaseBERT and BioBERT models separately, following the architecture proposed by Huang et al.~\cite{huang2015bidirectional}.

\paragraph*{Evaluation settings} 

We employed the span-level precision, recall, and F1 scores as evaluation metrics. For a prediction to be considered a true positive, it must satisfy two conditions: (1) the predicted span is identical to the truth entity, and (2) they should have the same entity type. To assess the statistical significance of our models, we performed document-level bootstrapping for five BERT-based models. Specifically, for the test set, we sampled 367 documents with replacement and evaluated the model on these documents. This process was repeated 100 times, yielding a distribution of the performance metric, such as Kappa. From this distribution, we reported the 95\% confidence intervals.

\subsection{Developing the Large Language Models}

We included results from two LLMs in our study: GPT-4~\cite{openai2024gpt4} and the open-source Llama 3 model~\cite{llama3modelcard}. We applied one-shot prompting (Box \ref{box:one-shot}) and RAG-based prompting~\cite{lewis2021retrievalaugmentedgenerationknowledgeintensivenlp}. For the Llama 3 model, we tested both the small-scale 8B and the large-scale 70B variants. 

\begin{prompt}[One-shot prompting]
\label{box:one-shot}
\begin{spverbatim}
Carefully read the following sentence. Output any mention that can be classified as DRUG, EFFECT, or DOSE and return the mention in the format of a "token" - label. If there's a duplicate, return one is enough. 

A DRUG is defined as a drug, vitamin, or herb, but not a neurotransmitter. Here is an example for DRUG: 'Also, in addition to a saturated solution in terms of dissolved APAP, a cloudy solution has all that plus a shitload of suspended / unsettled APAP as well ... what you see at the bottom is indeed some APAP, but also a shitload of insoluble excipients/fillers/colorants.' Should return an output: "APAP" - DRUG.

An EFFECT is defined as a change in physical or mental state associated with the substance. An example for EFFECT: `Is there anything anyone can recommend for the RLS ?' should return an output: 'RLS' - EFFECT;

A DOSE is defined as the quantity of a medicine or drug taken or recommended to be taken at a particular time. An example for DOSE: `I was thinking like 2 mg' should return an output: "2" - DOSE; If there are multiple mentions of DRUG and EFFECT in the input, output them all.

If there is no mention, please output: NO MENTION
\end{spverbatim}
\end{prompt}

We constructed a corpus by reshaping the training corpus for BERT-based models into JSON format, with each item consisting of a sentence and its relevant labels. With each retrieval round, the system utilized the BertSimilarity module from~\cite{xu} to identify the top 5 examples from the corpus most similar to the current query and append them to the RAG prompt (Box \ref{box:rag}). The concatenated prompt was then later fed into the generation pipeline. We provided an example of our final prompt, along with the query (extracted from the test dataset) and relevant examples, below.

\begin{prompt}[RAG prompting]
\label{box:rag}
\begin{spverbatim}
Carefully read the following sentence and output any mention that can be classified as DRUG, EFFECT, or DOSE and return the mention in a list of "token" - label. No justification is needed. 
     If there's no mention, return an empty list. 
     If there's a duplicate, return one is enough. Examples of similar sentences are provided as well.

Examples of similar sentences:

1) ... "example sentence": "Because you are wearing tracks deep into your neural processes by flooding them with cheap dopamine and serotonin frequently .", "example label": "[(dopamine, DRUG), (serotonin, DRUG)]" .... 
2)  ... "example sentence": "Still on an antidepressant but not one affecting serotonin as much.", ``example label": "[(antidepressant, DRUG), (serotonin, DRUG)]" ...
3) ... "example sentence": "I 2nd this ; your depression is reduced by dopamine stimulated by mu - receptor  temporarily .", "example label": ``[(dopamine, DRUG)]"
4) ... "example sentence": "I definitely get a burst of energy / dopamine from subs I've been on them for many years same dose ( actually less is more for me ) .", "example label": "[(dopamine, DRUG)]" 
5) ... "example sentence": "Benzos just do nt touch me antidepressants do not help and stimulants are a no go." ... 

Input sentence:
But I don't know of any antidepressants that are selective dopamine reuptake inhibitors ...
\end{spverbatim}
\end{prompt}

\paragraph*{Evaluation settings} 

Given that models typically struggle to correctly offset tokens due to insensitivity to numbers, we used document-level precision, recall, and F1 scores as our evaluation metrics. For a prediction to be classified as a true positive, it must meet two criteria: (1) the predicted entity must exactly match the true entity within each document, and (2) the predicted entity and the true entity must share the same entity type. If an entity appears multiple times, it was counted only once per document. Note that although our prompt included an instruction to extract DOSE, the subsequent scoring process became highly complex and subjective. The main reason is that DOSE entities are typically composed of multiple tokens, and it is common for LLMs to extract parts of them or return them as separate pairs, which makes evaluation difficult. Thus, we did not include the DOSE in the LLM report, and F1 scores were calculated without considering the DOSE. 

\section{Results}
\subsection{Dataset Records}
\begin{figure}
\centering
\includegraphics[width=.5\columnwidth]{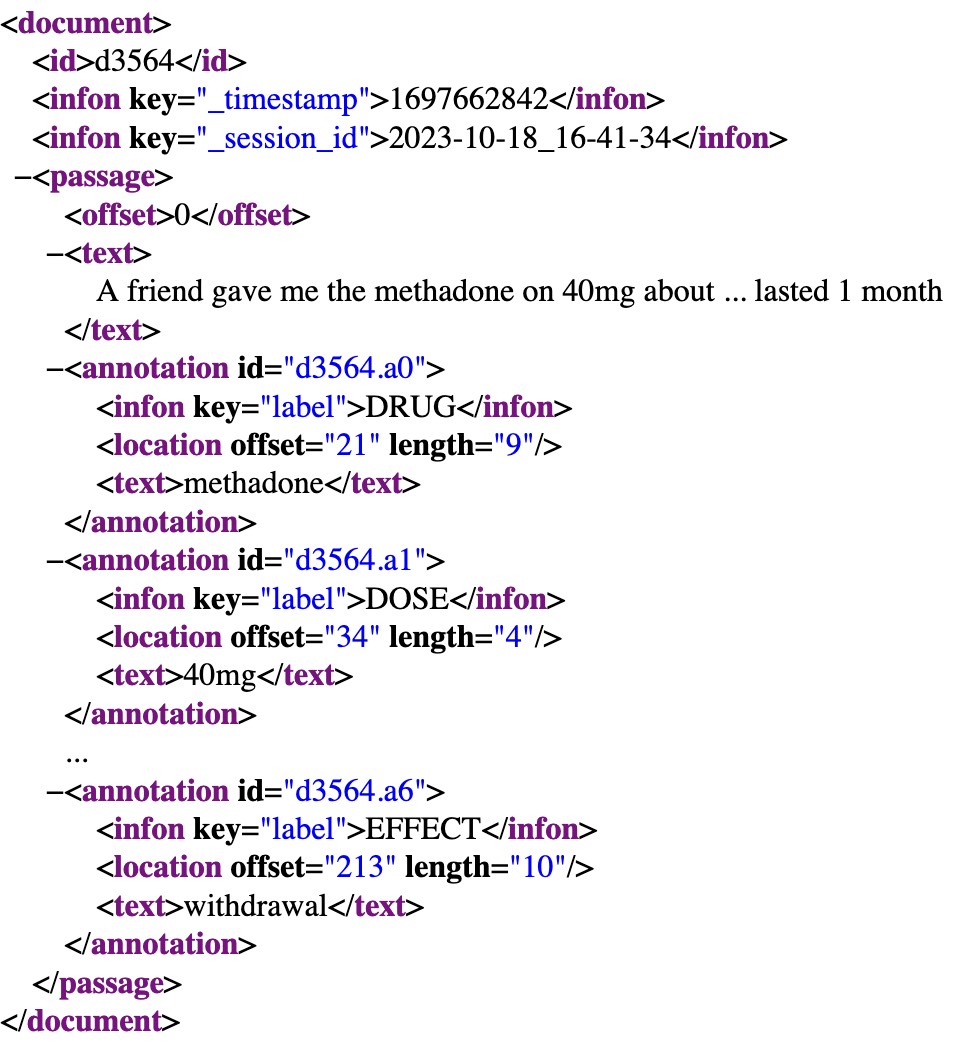}
\caption{BioC format.}
\label{fig:three sin x}
\end{figure}

\begin{figure}
\centering
\includegraphics[width=.3\columnwidth]{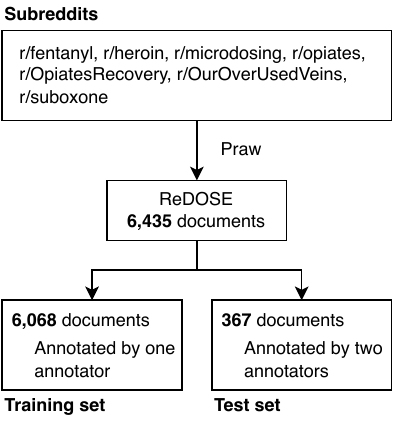}
\caption{Creation of the \redose dataset.}
\label{fig:workflow}
\end{figure}

\redose contains 6,435 documents and 15,469 sentences with 6,267 drug-related entities. There are 4,784 DRUG, 750 EFFECT, and 733 DOSE entities.
The dataset is split into 6,068 documents for training and 367 documents for testing (Figure \ref{fig:workflow}). At the sentence level, it is split into 14,260 sentences for training and 1,209 for testing. Table \ref{tab:redose} outlines the entity distribution: the training set includes 4,004 DRUG, 647 DOSE, and 674 EFFECT entities, while the test set contains 833 DRUG, 98 DOSE, and 128 EFFECT entities. While annotations can be represented in various formats, we used the BioC XML format due to several considerations~\cite{comeau2013bioc-o}, especially because the format is simple and easy to modify, allowing analysis tools to be applied rapidly. A sample XML is provided; the starting point for the BioC format is a \texttt{collection} of documents (Figure~\ref{fig:three sin x}).

Each \texttt{document} consists of a series of passages. Every \texttt{passage} includes an \texttt{offset}, which suggests the character offset within the parent document, and \texttt{text}, which stores the actual text of the passage. In each passage, annotations, which identify entities of DRUG, DOSE, and EFFECT, are applied directly to the surface \texttt{text}, which means the exact words as they appear rather than the transformed or normalized clinical concepts. The annotation contains \texttt{location}; it specifies the starting \texttt{offset} and the \texttt{length} of the annotated text within the passage. Besides, the IAA between our two annotators on the validation dataset is 75.1$\%$. The labels from both annotators are all included in our dataset in the testing folder.
\begin{table}[t]
\caption{Description of \redose.}
\label{tab:redose}
\centering
\small
\begin{tabular}{lrrr}
\toprule
     &  {Training} & {Test} & {Total}\\
\midrule
Document & 6,068 & 367 & 6,435\\
Sentence & 14,260 & 1,209 & 15,469\\
Entity\\
~~DRUG  & 4,004 & 833 & 4,784\\
~~EFFECT & 674 & 128 & 750\\
~~DOSE & 647 & 98 & 733\\
\bottomrule
\end{tabular}
\end{table}
\subsection{Performance of the BERT-based models}

Table~\ref{tab:entity results} shows the performance for different BERT-based models in entity extraction tasks.
BaseBERT achieved the highest recall for DRUG extraction (0.873), compared with BioBERT (0.838) and BiomedBERT (0.787). Additionally, it achieved a competitive F1-score of 0.882. In comparison, BioBERT achieved the highest recall for DOSE extraction (0.788), compared with BaseBERT (0.752) and BiomedBERT (0.714). Notably, BiomedBERT achieved the highest precision in DRUG extraction (0.907), compared with BaseBERT (0.891) and BioBERT (0.892).
However, the analysis of these metrics also highlighted significant variability in EFFECT extraction. Specifically, BiomedBERT showed a considerably low recall of 0.025 and an F1-score of 0.043, suggesting that while this model is robust for DRUG extraction, it struggles to detect more nuanced entities, such as EFFECT. For BaseBERT and BioBERT, the performance is neither competitive with EFFECT; they show a precision of 0.424 and 0.380, a recall of 0.231 and 0.167, and an F1-score of 0.298 and 0.230, respectively. To determine if better performance could be achieved, we added CRF+LSTM layers to the BaseBERT and BioBERT models.

\begin{table}[t]
\caption{Performance Comparison for BERT-based models.}
\label{tab:entity results}
\centering
\small
\begin{tabular}{lc@{~~~~}c@{~~~~}c}
\toprule
& {Precision} & {Recall} & {F1-score} \\
\midrule
\multicolumn{4}{l}{BaseBERT}\\
~~DRUG & 0.891 (0.889, 0.893) & \textbf{0.873} (0.871, 0.875) & \textbf{0.882} (0.880, 0.884) \\
~~EFFECT & 0.424 (0.411, 0.437) & \textbf{0.231} (0.222, 0.240) & \textbf{0.298} (0.288, 0.308) \\
~~DOSE & \textbf{0.649} (0.640, 0.658) & 0.752 (0.742, 0.761) & \textbf{0.696} (0.688, 0.705) \\
~~\textit{micro avg} & 0.843 (0.840, 0.845) & 0.810 (0.808, 0.812) & 0.826 (0.824, 0.829) \\
\midrule
\multicolumn{4}{l}{BioBERT}\\
~~DRUG & {0.892 (0.890, 0.894)} & 0.838 (0.835, 0.840) & 0.864 (0.862, 0.866)\\
~~EFFECT & 0.380 (0.364, 0.395)& 0.167 (0.158, 0.175) & 0.230 (0.220, 0.241)\\
~~DOSE & 0.620 (0.612, 0.628) & \textbf{0.788} (0.779, 0.797) &  0.694 (0.685, 0.702)\\
~~\textit{micro avg} & 0.838 (0.836, 0.840) & 0.779 (0.776, 0.781) & 0.807 (0.805, 0.810) \\

\midrule
\multicolumn{4}{l}{BiomedBERT}\\
~~DRUG & \textbf{0.907} (0.905, 0.909) & 0.787 (0.784, 0.790) & 0.843 (0.841, 0.845) \\
~~EFFECT & 0.178 (0.153, 0.202) & 0.025 (0.022, 0.028) & 0.043 (0.038, 0.048) \\
~~DOSE & 0.636 (0.627, 0.645) & 0.714 (0.705, 0.723) & 0.672 (0.664, 0.680) \\
~~\textit{micro avg} & 0.863 (0.860, 0.865) & 0.718 (0.716, 0.721) & 0.784 (0.782, 0.787) \\

\midrule
\multicolumn{4}{l}{BaseBERT+CRF+LSTM}\\
~~DRUG & 0.846 (0.844, 0.849) & 0.871 (0.868, 0.873) & 0.858 (0.856, 0.860) \\
~~EFFECT & 0.261 (0.252, 0.270) & 0.208 (0.200, 0.217) & 0.231 (0.222, 0.239) \\
~~DOSE & {0.633 (0.624, 0.642)} & 0.719 (0.710, 0.727) & {0.673 (0.665, 0.681)} \\
~~\textit{micro avg} & 0.787 (0.785, 0.790) & 0.802 (0.800, 0.804) & 0.792 (0.793, 0.797) \\
\midrule
\multicolumn{4}{l}{BioBERT+CRF+LSTM}\\
~~DRUG & 0.883 (0.881, 0.885) & {0.868 (0.866, 0.871)} & 0.876 (0.874, 0.877) \\
~~EFFECT & \textbf{0.497} (0.483, 0.511) & 0.208 (0.199, 0.217) & 0.291 (0.281, 0.301) \\
~~DOSE & 0.592 (0.584, 0.601) & {0.730 (0.721, 0.738)} & 0.653 (0.646, 0.661) \\
~~\textit{micro avg} & 0.836 (0.834, 0.838) & {0.801 (0.799, 0.804)} & 0.818 (0.816, 0.820)\\
\bottomrule
\end{tabular}

\end{table}

When comparing BERT models with the CRF+LSTM architecture, we found that the BioBERT+CRF+LSTM and BaseBERT+CRF+LSTM models performed better in identifying EFFECT than BiomedBERT, achieving notably higher F1-scores of 0.231 and 0.291, respectively, compared to 0.043. However, their performance in extracting DRUG and DOSE was comparable to that of other BERT-based models.

\subsection{Performance of the Large Language Models}

Table \ref{tab:llm} compares the performance of four types of LLMs in DRUG and EFFECT extraction. Among these, Llama-3 70B achieved the highest performance in DRUG extraction with a precision of 0.74, a recall of 0.84, and an F1-score of 0.79. In contrast, GPT-4 showed stronger performance in EFFECT extraction with a recall of 0.41. In contrast, GPT-4 exhibited more balanced performance across both entity types. Although its DRUG F1-score (0.72) was slightly lower than that of Llama-3 70B, GPT-4 achieved a much higher recall on EFFECT extraction (0.41), resulting in the best overall F1-score for EFFECT. Besides, GPT-4’s micro averages (precision: 0.46, recall: 0.69, F1-score: 0.55) indicated average-level performance but fell slightly behind Llama-3 70B (precision: 0.64, recall: 0.76, F1-score: 0.70).

Further experiments showed that the RAG approach significantly improved Llama-3 8B's performance in DRUG extraction, with a recall exceeding 80\% vs. 44\% without RAG and an F1-score 75\% vs. 55\% without RAG. However, gains in EFFECT extraction were modest, with the F1-score increasing only from 0.05 to 0.07.

\begin{table}[t]
\caption{Performance Comparison for LLMs.}
\label{tab:llm}
\centering
\small
\begin{tabular}{lccc}
\toprule
 & Precision & Recall & F1-score\\
\midrule
Llama-3 Instruct 8B\\
~~DRUG & 0.71 & 0.44 & 0.55\\
~~EFFECT & 0.03 & 0.11 & 0.05\\
~~\textit{micro avg} & 0.50 & 0.41 & 0.45\\
\midrule
Llama-3 Instruct 70B\\
~~DRUG & {\textbf{0.74}} & {\textbf{0.84}} & {\textbf{0.79}}  \\
~~EFFECT & {0.02} & {0.04} & {0.03}  \\
~~\textit{micro avg} & 0.64 & 0.76 & 0.70\\
\midrule
Llama-3 Instruct 8B + RAG\\
~~DRUG & 0.71 & 0.80 & 0.75 \\
~~EFFECT & 0.04 & 0.15 & 0.07\\
~~\textit{micro avg} & 0.50 & 0.75 & 0.60\\
\midrule
GPT-4\\
~~DRUG & {0.72} & {0.73} & {0.72} \\
~~EFFECT & {\textbf{0.07}} & {\textbf{0.41}} & {\textbf{0.12}} \\
~~\textit{micro avg} & 0.46&0.69&0.55\\
\bottomrule
\end{tabular}
\end{table}

\section{Discussion}

\redose is a novel dataset that not only integrates with other datasets to develop a comprehensive medical NER dataset but also provides valuable insights into substances used by the public and the vernacular terms they used, which often differ markedly from formal medical terminologies. For example, the drug ``fentanyl'' may be referred to as ``fent'' or ``f'' on online forums. Epidemiologists unfamiliar with these terms may miss emerging hotspots, delaying intervention. Physicians unfamiliar with informal terms might misunderstand a patient's pattern of substance use, or have difficulty building rapport. This highlights another crucial aspect of \redose -- it bridges the gap between the formal medical language used in healthcare settings and the colloquial terms used in everyday discussions about drugs. 

When testing both the small-scale 8B and the large-scale 70B variants for Llama 3, 
we observed that the base model demonstrated a significantly weaker ability to follow the instructions in the prompts. This observation is consistent with prior work demonstrating that instruction-tuned models outperform base models in the medical field~\cite{Hou2024,Wu2025}. These findings underscore the importance of instruction-tuning in instruction-intensive tasks, particularly in the medical field. 

Regarding BERT-based models, several factors may contribute to observed variations in performance. First, discussions on platforms like Reddit about drugs, dosages, and effects often utilize language more akin to everyday speech. This linguistic alignment might account for the similar performance levels observed in BaseBERT and BiomedBERT. 
Secondly, the relative simplicity of the task may limit the advantages of employing more complex models, such as the CRF+LSTM architecture, which has not shown significant improvements over simpler models.
Additionally, tokenization plays a crucial role in the performance of NER systems. For instance, both BaseBERT and BioBERT split the word ``chloride'' into tokens ``ch-lo-ride'', treating ``lo'' and ``ride'' as special tokens and leaving ``ch'' for prediction by the model. This approach to tokenization could significantly affect the performance. 
Moreover, there appears to be confusion in how BERT-based models manage certain terminology. For instance, the term ``Straight" is labeled as O (Outside) in the training set but is labeled as DRUG in the testing set due to different meanings represented in the sentence, which suggests inconsistencies in dataset annotations. Similarly, the term ``Oxy" is annotated as B-DRUG (Beginning of DRUG) in the training data but changes to I-DRUG (Inside of DRUG) in the testing set. 
Finally, the models struggle particularly with the EFFECT category. The annotation of EFFECT often requires recognizing spans of multiple words, such as in ``depress your respiration''. While the models may accurately identify ``depress'' as B-EFFECT (Beginning of EFFECT), they frequently fail to recognize ``your'' as part of the same entity (I-EFFECT), leading to incorrect or incomplete entity predictions.

In contrast, the performance differences among LLMs can be attributed to several other factors. Firstly, unlike BERT-based models, LLMs operate without supervision and therefore lack access to label distribution. This limitation leads to LLMs having a poor estimate of entity frequency in the dataset. Consequently, we observe more false-negative extractions from LLMs than from the gold-standard labels. Secondly, our evaluation applies a strict criterion: extracted entities are deemed incorrect unless every word in an entity is output exactly as annotated. 

This high standard particularly challenges EFFECT extraction, since one may use diverse language to describe their feelings. In our dataset, EFFECT spans range from canonical affective terms (e.g., “euphoria”) to abbreviations (e.g., “WD” for withdrawal, as well as shorthand such as “PWD”), and even metaphorical expressions (e.g., “kill your hormones”). Such variability increases boundary ambiguity and often necessitates co-reference resolution to determine the referent of an effect mention. Moreover, under our strict criteria, predictions that deviate by even a single token are counted as incorrect, thereby disproportionately penalizing EFFECT relative to more lexically stable entity types. A related challenge arises for DOSE extraction, as dosage information is frequently expressed in relative or ranged forms (e.g., proportions or intervals) rather than as standardized absolute quantities. Future work could improve EFFECT extraction by augmenting training data with abbreviation and slang normalization (e.g., mapping WD to withdrawal) and lexicon-assisted weak supervision, and by incorporating context modeling beyond sentence-level NER, such as co-reference resolution, to better link effects to the triggering drug-use context.

This study has several limitations. 
One limitation of \redose is the potential for selection bias. Not all substance users share their personal experiences in public forums or honestly represent their full experience.  \redose focuses on the dose-effect relationship, a relationship essential to understanding toxicity and problematic usage. But, dose-effect relationships do not tell the whole story of substance use. Medical records provide a more appropriate data source for analyzing fatalities. 
Our benchmarks evaluated the original large language models with a one-shot trial. Given the length of our dataset, we did not perform fine-tuning on \redose, which might improve performance. Finally, while annotations were curated by domain-informed annotators, future extensions of \redose may benefit from broader engagement with experts. We hope that future efforts will address these limitations and contribute to the development of improved models and benchmarks.

Notably, BERT-based models outperformed state-of-the-art large language models in terms of recall and F1-score. However, BaseBERT achieved higher recall and F1 scores compared to domain-specific models. This may be attributed to the nature of Reddit content, where user-generated text tends to align more closely with general language corpora than with strictly medical jargon. Additionally, attempts to enhance BERT-based models with CRF-LSTM layers did not yield better performance. This may be due to the increased complexity of these layers, which may have introduced noise rather than improving the model's ability to extract relevant information from unstructured text. These results suggest that simpler models may be better suited to user-generated content, particularly when the text deviates from domain-specific language.

Several directions remain for future exploration. First, expanding the annotation schema would improve coverage of colloquial expressions, rare substances, and nuanced descriptions of effects, which are currently underrepresented. Such schema expansion may also facilitate the identification of emerging or previously unseen drugs by capturing recurring patterns of drug use discourse, even when specific substance names are not yet standardized or widely known. Second, more advanced fine-tuning approaches, including parameter-efficient tuning and domain-adaptive pretraining, could further enhance model robustness across entity types. Beyond the current English-only dataset, cross-lingual extensions of \redose would enable monitoring of substance use discourse in diverse communities worldwide. Another promising direction is multimodal integration, incorporating non-textual cues such as emojis, images, or memes may enrich the contextual interpretation of user posts. 

\section{Conclusion}
To bridge the gap between formal medical terminology and the colloquial language commonly used in online discussions about drug use, we proposed \redose. By curating a dataset of 6,435 Reddit posts with annotated drug names, doses, and effects, \redose enabled the development and benchmarking of machine learning models to extract clinically relevant information from unstructured social media text. The benchmarking results demonstrated that while LLMs such as GPT-4 and Llama-3 showed promise, domain-specific, fine-tuned models, such as BioBERT and BiomedBERT, also achieved high precision, particularly for drug extraction. The results suggest that RAG may be a better alternative to manually selecting the most representative examples under this setting. \redose not only provides a valuable resource for advancing natural language processing in the substance use domain but also emphasizes the importance of integrating user-generated data into public health research to detect emerging substance use trends and guide interventions. Future work should address dataset biases, refine annotations, and explore fine-tuning methods to further improve model performance.

\section*{Acknowledgment}

This research was supported by the National Library of Medicine under the grant numbers R01LM014306.

\bibliographystyle{unsrtnat}
\bibliography{citation}

\end{document}